\documentclass[conference]{IEEEtran}
\IEEEoverridecommandlockouts
\usepackage{cite}
\usepackage{amsmath,amssymb,amsfonts}
\usepackage{algorithmic}
\usepackage{graphicx}
\usepackage{textcomp}
\usepackage{xcolor}
\usepackage{url}
\usepackage{longtable}
\usepackage{verbatim}
\usepackage{multirow}
\usepackage{pdfpages}

\begin{document}

\title{Research Challenges and Progress in Robotic Grasping and Manipulation Competitions}
%Suggestion for title: Research Challenges and Progress in the Robotic Grasping and Manipulation Competitions (RGMC)

\author{Yu Sun, Joe Falco, M\'{a}ximo A. Roa, and Berk Calli% <-this % stops a space
\thanks{Accepted for publication by IEEE Robotics and Automation Letters.}%Use only for final RAL version
\thanks{Yu Sun is with the University of South Florida, Joe Falco is with the National Institute of Standards and Technology (NIST), M\'{a}ximo A. Roa is with the German Aerospace Center (DLR), and Berk Calli is with Worcester Polytechnic Institute (WPI). E-mails: yusun@usf.edu, falco@nist.gov, maximo.roa@dlr.de, bcalli@wpi.edu}
}

% make the title area
\maketitle

\begin{abstract}
This paper discusses recent research progress in robotic grasping and manipulation in the light of the latest Robotic Grasping and Manipulation Competitions (RGMCs). We first provide an overview of past benchmarks and competitions related to the robotics manipulation field. Then, we discuss the methodology behind designing the manipulation tasks in RGMCs. We provide a detailed analysis of key challenges for each task and identify the most difficult aspects based on the competing teams' performance in recent years. We believe that such an analysis is insightful to determine the future research directions for the robotic manipulation domain.
\end{abstract}


\maketitle

\section{Introduction}
\IEEEPARstart{R}{obotic} grasping and manipulation as a research field had tremendous growth in the last decade.  Researchers have made significant progress in different areas that prevented robots from reliably handling household items, mechanical parts, and packages. The progress in robotic grasping and manipulation has shown new application promises that led to a renewed interest in robotics from the general public, industry, and government agencies.
Nevertheless, the growth and progress have not been even. Some challenges receive a great deal of well-deserved attention because they are either obvious or standing in the way of big commercialization potential. Some challenges might not be as popular and remain unsolved for decades, but they could be crucial for many applications. Some challenges may have changed or vanished because a new kind of hardware becomes available or an engineering solution became adequate. 
A comprehensive overview of the major challenges not only helps us analyze the history of the robotic grasping and manipulation field, but also allows us to determine future research directions. We should keep track of the challenges, their changes, and the progress made to solve them. Competitions rooted in real-life applications could be an ideal vehicle for this purpose.  

%This is already in the background
%In the last decades, several well-attended robotic grasping and manipulation challenges/competitions have been organized with different focuses. Robotic Grasping and Manipulation Competition (RGMC) \cite{graspcompetit} have been organized four times at IROS; Amazon Picking Challenge \cite{Eppner17} is the most well-funded event focusing on warehouse robotics applications; RoboCup@Home \cite{robocup-home} is a long standing competition with a service robotics focus. All these competitions, among others, foster innovation in the field, highlighted the gaps in the manipulation research, and was influential to connect the research outcomes in academia to the needs in industry.

%Separate challenges that need systematic research solutions from the ones needs engineering solutions. 
This paper discusses recent progress in the robotic manipulation field based on the recent Robotic Grasping and Manipulation Competitions (RGMC)\footnote{Certain commercial entities and items are identified in this paper to foster understanding.  Such identification does not imply recommendation or endorsement by the National Institute of Standards and Technology, nor does it imply that the materials or equipment identified are necessarily the best available for the purpose.}. For defining the RGMCs, we designed over 40 grasping and manipulation tasks that reflect realistic scenarios in service robotics and industry.  Each task has a detailed setup and requirement description and scoring rules. The task details can be found on the competition webpages \cite{rgmc2016, rgmc2017, rgmc2019, rgmc2020} and \cite{sun2016robotic,falco2016robotic}. This paper presents the challenges and the progress of the competing teams, identifies critical areas preventing better performance in robotic manipulation, and provides observations regarding future research directions.

%This paper focuses on grasping approaches presented in the competitions. Due to the limitation of the competition, the approaches are also limited by the short-period of preparation time, limited resources and participations. 

%The grasping approached in the competitions favors technologies that are easy to implement, robust. 

%Convert benchmarks to competitions

%%%%%%%%%%%%%%%%%%%%%%%%%%%%%%%%%%%%%%%%%%%%%%%%%%%%%%%%%%%%%%%%%%%%%%
\section{Background}

\subsection{Related Robotic Grasping and Manipulation Benchmarks} 
Establishing experimentation methodologies that allow comparison across different research groups is still a pending challenge in robotics. The large variety in robotic platforms, setups, and software implementations poses numerous difficulties to achieve common experiment protocols.
Even the pressure to publish plays a role in this lack of comparison, as applying the benchmarking protocols is a thorough and time-consuming process, which is sometimes neglected under the pressure of publication deadlines. In this sense, initiatives such as the Reproducible Articles (R-articles) of the IEEE Robotics \& Automation Magazine (RAM) \cite{bonsignorio2017new} aim to encourage easy reproduction of results via adopting and/or presenting detailed experiment protocols. Several initiatives of benchmarks for grasping and manipulation have been proposed, focusing on different levels of the manipulation system:

\begin{table*}[t]
    \centering
        \caption{Robotic competitions involving manipulation}
    \label{tab:competitions}
\begin{tabular}{p{0.2\textwidth}|p{0.08\textwidth}|c|c|c|c|c|c|p{0.4\textwidth}}
\hline
    Competition/Challenge & Years & \multicolumn{2}{c|}{Tasks*} & \multicolumn{4}{c|}{Focus**}  & Description\\
      &  & Fi & Fr & Gr & Ma & As & Mm &\\
    \hline \hline
     RoboCup@Home \cite{robocup-home} & 2006-2021 & X & X & X & X & & X & Domestic service tasks, different fixed challenges each year\\
        \hline
     DARPA Robotics Challenge \cite{darparc} & 2012-2015 & X & & X & X &  & X & Disaster response scenarios\\
         \hline
    RoCKIn \cite{rockin} & 2014, 2015 & X & & X & X & & X & Two tracks: RoCKIn@Home (domestic environment) and  RoCKIn@Work (factory environment)\\
        \hline
     European Robotic Challenge (EuRoC) \cite{euroc} & 2014-2017 & & X & X & X & & X & Reconfigurable manufacturing cells, shop floor logistics and manipulation, and plant servicing and inspection\\
     \hline
    Amazon Picking Challenge (APC) \cite{Eppner17} & 2015-2016 & X & & X & X & & & Logistics scenario, retrieving items\\
         \hline
    IROS Robotic grasping and Manipulation Competition (RGMC)  \cite{sun2016robotic,falco2016robotic} & 2016-2017, 2019-2020 & X & & X & X & X & & Different tracks: service, manufacturing, and logistics\\
    \hline
     Cybathlon arm prosthesis race \cite{cybathlon} & 2016, 2020& X & & X & X & & & Solution of everyday tasks using arm prostheses\\
         \hline
     Amazon Robotics Challenge \cite{arc} & 2017 &X & & X & X & & & Logistic environment, as evolution of APC \cite{Eppner17}, pick and stow items\\
         \hline
     ICRA Mobile Manipulation Challenge  \cite{icrammc17},\cite{icrammc19} & 2017, 2019 & X & & X & & & X & Navigate, pick and place items\\
    \hline
     World Robot Challenge (WRC) \cite{wrs} & 2018, 2021 & X & & X & X & X & & Three categories: industrial scenarios (agile manufacturing), service robotics (home and convenience stores), and disaster robotics (inspection and maintenance in a plant)\\
     \hline
     IROS Fan Robotic Challenge \cite{irosfan} & 2018 & X & & X & X & & & Pick up and manipulate a Spanish fan\\
        \hline
    IROS Mobile Manipulation Hackathon \cite{irosmmh} & 2018 & & X & X & X & & X & Freely-chosen application to showcase both mobility and manipulation\\
    \hline
    RoboSoft Competition \cite{robosoft}& 2018-2021 & X & & X & X & & & Industrial, surgical or domestic scenarios in the manipulation track\\
         \hline
     Smart City Robotics Challenge (SciRoc) \cite{sciroc} & 2019, 2021 & X & & X & X & & & Different episodes, or challenges, in domestic and logistic scenarios\\
         \hline
    OCRTOC, Open Cloud Robot Organization Challenge \cite{9619915} & 2020, 2021 & X & & X & X & & & Table reorganization problem, tested in a remote lab\\
         \hline
     Real Robot Challenge \cite{rrc} & 2020-2021 &X & & X & X & & & Grasping and in-hand manipulation tasks in a remote platform\\
         \hline
\multicolumn{9}{l}{*Tasks: Fi: Fixed, Fr: Free}\\
\multicolumn{9}{l}{**Focus: Gr: Grasping, Ma: Manipulation, As: Assembly, Mm: Mobile manipulation}
\end{tabular}
\end{table*}

\noindent 
\underline{Mechanism level:} In these types of benchmarks, the intrinsic capabilities of the mechanisms (often times end-effectors) are considered and measured. A benchmarking protocol for assessing and comparing the grasping abilities of robotic hands is presented in \cite{llop2019anthropomorphic}. The benchmark in \cite{negrello2020benchmarking} assesses the robustness and resilience of the robotic hands by determining the impulsive conditions that break their grasp and their mechanism itself. A detailed list of procedures for analyzing the mechanical properties of the robotic hands is given in \cite{falco2018performance}. A specific benchmark for compliant hands is provided in \cite{Sotiropoulos2018}.

\noindent
\underline{Algorithm level:} These benchmarks evaluate the performance of a specific algorithm in the robotic manipulation pipeline. The benchmarks in \cite{bekiroglu2019benchmarking,bottarel2020graspa} assess the performance of grasp planning algorithms. In \cite{falco2016benchmarking}, the force control capabilities of the system are analyzed. A platform-independent method for quantifying the motion planning performance is presented in \cite{lagriffoul2018platform}. Object segmentation and pose estimation data sets and benchmarks are commonly used in the robotics and computer vision community, with recent examples in \cite{xu2018youtube,grenzdorffer2020ycb}.

\noindent
\underline{System level:} These benchmarks consider the task performance of a robotic system as a whole, fully integrated autonomous solution (perception, planning, control). The box and blocks test in \cite{morgan2019benchmarking} assesses the pick-and-place performance of a robotic system. Similarly, pick-and-place abilities in logistics scenarios are evaluated in \cite{Mnyusiwalla20}. Inspired by the Amazon Picking Challenge, a shelf picking benchmark is presented in \cite{Leitner2017}. In-hand manipulation performance is quantified in \cite{cruciani2020benchmarking}. Assessment for various challenging manipulation tasks such as cloth \cite{garcia2020benchmarking}, bimanual \cite{chatzilygeroudis2020benchmark} and aerial \cite{suarez2020benchmarks} are also provided as system-level benchmarks.

\subsection{Related Competitions and Challenges}

The robotics community has a long history in organizing competitions and challenges.  
Over the last decade, there have been multiple competitions that involve robotic manipulation in different degrees of complexity, pushing forward the research in the field.
The competitions focus on different aspects, such as pure grasping, manipulation, assembly, or even mobile manipulation, as summarized in Table~\ref{tab:competitions}. In terms of tasks, those competitions adopt two approaches:
\begin{itemize}
    \item Completing a fixed set of tasks: the environment, objects, and rules are defined in advance. Thus, the tasks provide an objective measure of progress intra-edition, i.e., comparison of performance of different teams, and inter-edition, i.e., measuring progress across different years on the same set of tasks. The disadvantage of this approach is that teams usually over-engineer their solution (or work around the rules) to fulfill the intended task.
    \item Demonstration of a freely-chosen application: this open format allows the teams to demonstrate their strengths in a self-selected scenario, usually following very general constraints, e.g., using a common robotic platform or demonstrating tasks that include prescribed components. The disadvantage is that demonstrations are hardly comparable among them.
\end{itemize}

The Robotic Grasping and Manipulation Challenge (RGMC) uses system level benchmarks and defines a fixed set of tasks in each track to be solved using a robot manipulator. The descriptions of the tasks are provided in the next section.

%%%%%%%%%%%%%%%%%%%%%%%%%%%%%%%%%%%%%%%%%%%%%%%%%%%%%%%%%%%%%%%%%%%%%%
\section{Task Designs}
To evaluate robotic systems' capability, we have designed over 40 real-life tasks based on our knowledge of the research challenges in robotic grasping and manipulation.  All these tasks have detailed protocols, rules, and scoring policies. For instance, the task descriptions and rules of the 2016 and 2017 RGMCs can be found in \cite{sun2016robotic}. Each competition comprises challenges that span a large set of robotic manipulation capabilities with varying difficulty levels. Easier tasks are expected to be solved by most teams, while some challenging tasks are not expected to be fully accomplished by any of the teams. Such variety in difficulty serves multiple purposes: Easier tasks encourage participation and constitute a starting point for new robotics researchers. Challenging tasks aim to differentiate the most successful teams while encouraging robot/algorithm design and integration innovation. Limited by natural constraints of the competition settings (e.g., timeline, space, and equipment), the tasks are carefully composed to appeal to a large group of researchers and research interests. Therefore, some of the tasks that were selected in a certain year might be replaced by others in the following years, even though the teams did not fully accomplish them.  

Here we present the tasks in the RGMCs according to the research challenges associated with them.  There are plenty of unsolved research challenges in perception, grasping, manipulation, and mechanical design trade-offs for both real-time service tasks as well as for manufacturing/assembly tasks \cite{8957300}. 

\subsection{Research Challenges in Perception}
Perception is critical to most robotic grasping and manipulation tasks. A significant amount of work has been dedicated to solving various challenges in robotic perception and, over the years, we have seen great improvement, especially after depth cameras became widely accessible and high-precision tactile sensors became less expensive. However, many challenges remain unsolved, and while designing our competitions we highlight these challenges to gauge progress.

\begin{figure}[t]
	\centering
	\includegraphics[width=\columnwidth]{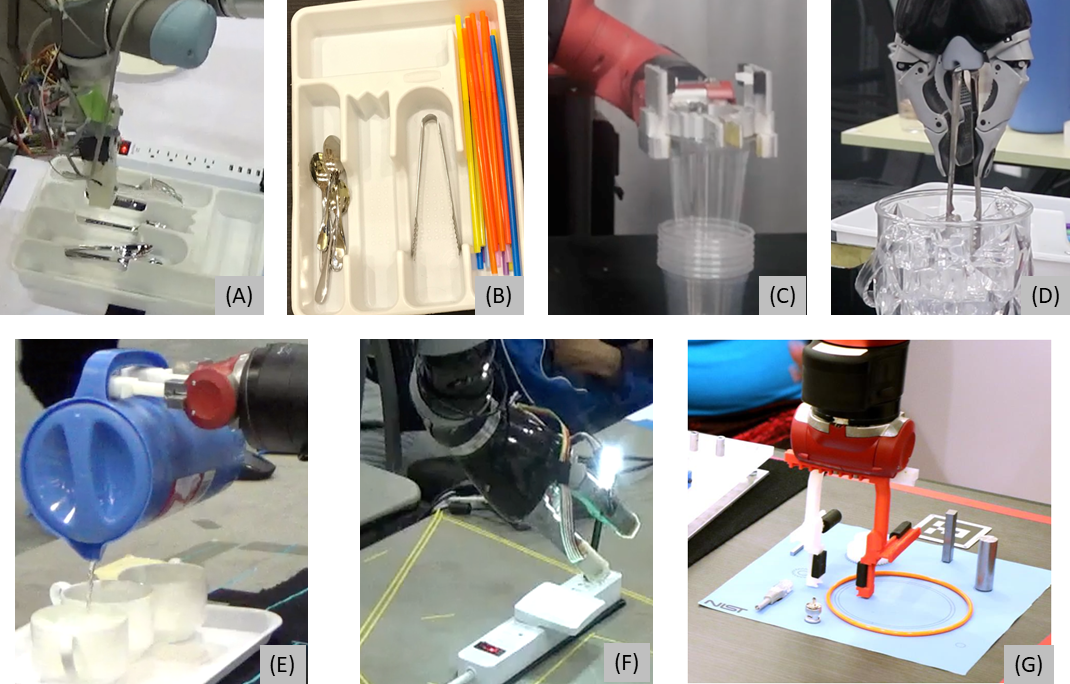}
%    \includegraphics[height=0.13\textwidth]{figures/silverware-2017.png}
%    \includegraphics[height=0.13\textwidth]{figures/tray-2019.png}	%\includegraphics[height=0.13\textwidth]{figures/com-2019-serev-pickcup.png}
%    \includegraphics[height=0.13\textwidth]{figures/com-2019-serev-icecube.png}	
%    (A)~~~~~~~~~~~~~~(B)~~~~~~~~~~~~~(C)~~~~~~~~~~~~~(D)\\
%	
%	\includegraphics[height=0.155\textwidth]{figures/pour-2017.png}
%    \includegraphics[height=0.15\textwidth]{figures/straw-2019.png}
%	\includegraphics[height=0.155\textwidth]{figures/plug-2017.png}
%	\includegraphics[height=0.155\textwidth]{figures/belt_perception.png}
%	(E)~~~~~~~~~~~~~~~(F)~~~~~~~~~~~~~~~(G)\\
\caption{Task setups based on perception challenges. Description of the tasks is provided in Table~\ref{tab:perception-challenges}.}
	\label{fig:perception-challenges}
\end{figure}

\subsubsection{Objects with shiny surfaces} 
Computer vision has a long-standing difficulty in dealing with shiny objects. Because of the reflection, some shiny surface areas cannot be well-perceived by vision sensors. The missing areas make segmentation and pose estimation difficult, and drop the success rate of grasping and manipulation algorithms. In RGMCs, we have several objects with shiny surfaces, such as the silverware sets, 2017 RGMC (Fig. \ref{fig:perception-challenges}a) and spoons and tongs, 2019/2020 RGMCs (Fig. \ref{fig:perception-challenges}b). Also, parts such as the plastic assembly base (task board) and metallic components in manufacturing tasks (Fig.~\ref{fig:perception-challenges}g) can present perception difficulties.

\begin{table}
    \centering
     \caption{Tasks reflecting perception challenges}
    \label{tab:perception-challenges}
    \begin{tabular}{p{0.13\textwidth}|p{0.3\textwidth}}
        \hline
        Research challenges & Tasks \\
        \hline \hline
        Shiny objects 
        & -Pick up silverware (Fig.~\ref{fig:perception-challenges}a) \\
        & -Pick up a a polished spoon for stirring (Fig.~\ref{fig:perception-challenges}b)\\
        & -Pick up metal tongs to get ice cubes (Fig.~\ref{fig:perception-challenges}b)\\
        & -Pick up metal pegs for insertion \\
        & -Locate insertion holes on task board \\
            \hline
        Translucent objects 
        & -Pick up transparent cup lid (Fig.~\ref{fig:perception-challenges}c) \\
        & -Pick up ice cubes 
        (Fig.~\ref{fig:perception-challenges}d) \\
        & -Pour a certain amount of water into a cup (Fig.~\ref{fig:perception-challenges}e)\\
            \hline
        Precision/accuracy
        & -Insert electrical connectors (Fig.~\ref{fig:perception-challenges}f)\\  
        & -Insert pegs, fasteners 
        (Fig.~\ref{fig:grasp-challenges}f)\\
        & -Insert  gears
        (Fig.~\ref{fig:manipulation-challenges}b)\\
        & -Insert/route belts (Fig.~\ref{fig:perception-challenges}g)\\
%       & insert straw into a cup through its lid \\
            \hline
        \end{tabular}
\end{table}

\subsubsection{Translucent or transparent objects}
Translucent objects are also very challenging for robotic vision. In all four RGMCs, we have translucent to-go cups and lids (Fig.~\ref{fig:perception-challenges}c). We have also introduced a pile of ice cubes for a picking task in the last two RGMCs.  In general, the sensing aspect of the ice cube picking task is more challenging than the sensing for picking up a to-go cup and a lid, since the to-go cups and lids are standalone objects on the table, and they occupy a significant space to approximate their pose. On the other hand, the ice cubes are randomly piled up in an ice bucket (Fig.~\ref{fig:perception-challenges}d), and determining the pose of an individual ice cube among other (also translucent) ice-cubes within a translucent ice bucket is a very challenging task. In 2017 RGMC, the competition requested to pour a certain amount of water into a cup. Estimating the amount of water in the cup also provides a significant challenge since it is transparent (Fig.~\ref{fig:perception-challenges}e). There are currently no translucent or transparent objects used for the manufacturing tasks.

\subsubsection{Tasks requiring high precision or accuracy}
Tasks with tight tolerances usually require the perception part of the system to be precise. For typical peg-in-hole problems such as plugging a Universal Serial Bus (USB) light into a socket (Fig.~\ref{fig:perception-challenges}f), a robot would need to use vision to precisely localize the socket. To compensate for the inaccuracies of the localization, an exploration algorithm or a failure recovery approach is often times needed. Using force sensors to guide corrective motions is a common example of such exploration strategies.

\subsubsection{Other challenges} Service track objects are provided to teams prior to competitions depending on type and difficulty.  This time can range from hours to days. The teams usually utilize this time to collect data and learn models for object identification, segmentation, and planning. While developing techniques to model the objects in a short amount of time under a semi-controlled environment is a challenge, obtaining prior models of the objects significantly reduces the perception and planning difficulties of the task. If the objects are unknown (no learning data or models), the tasks could be much more difficult. However, for real service applications, learning object models is typically not feasible. In the case of the manufacturing track, design data is typically known, so all components and Computer Aided Design (CAD) data are provided to teams weeks in advance to the competition.  At the start of the competition all teams are given updated CAD data that reflects changes in the challenge tasks from the practice tasks, which imitates challenges associated with product changes in batch type production runs.

Table \ref{tab:perception-challenges} summarizes the tested perception challenges and the tasks reflecting them in the RGMCs.

\subsection{Research Challenges in Grasping}

\begin{figure}[t]
	\centering
    \includegraphics[width=\columnwidth]{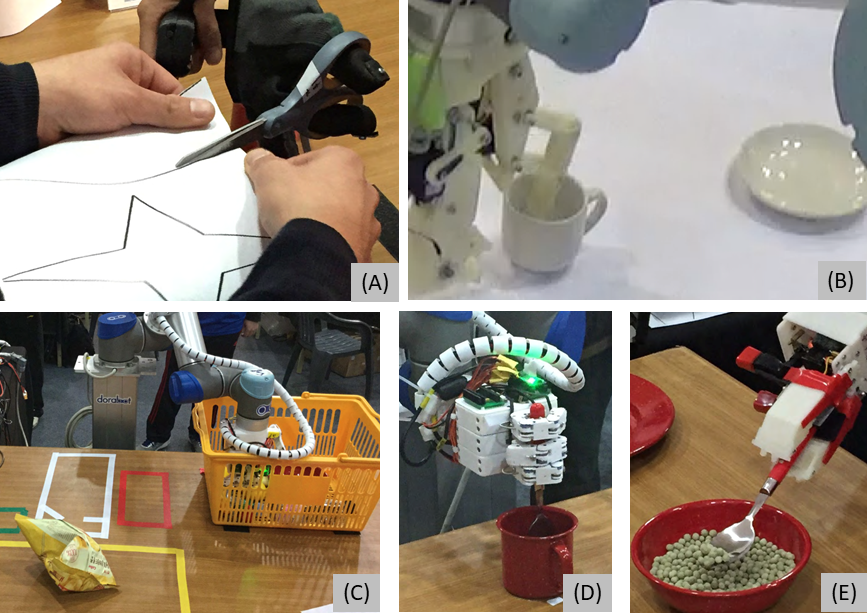}

    \vspace{1mm}
	\includegraphics[width=\columnwidth]{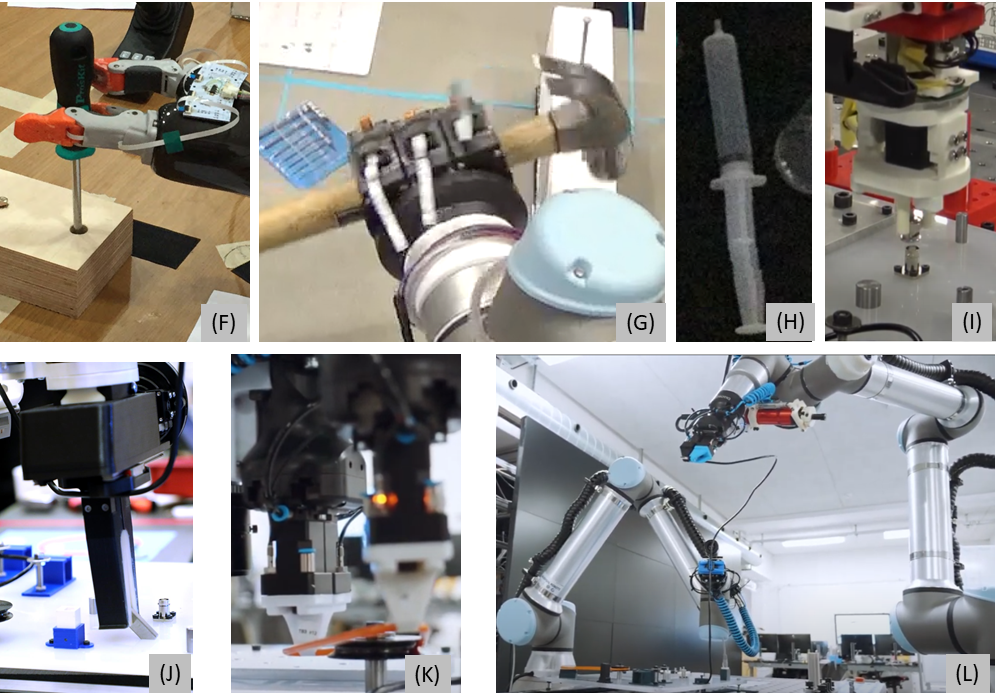}
%	\includegraphics[height=0.165\textwidth]{figures/cut-2016.png}
%	\includegraphics[height=0.165\textwidth]{figures/cup-saucer-2017.png}
%    (A)~~~~~~~~~~~~~~~~~~~~~(B)\\
%	\includegraphics[height=0.17\textwidth]{figures/pick-place-2016.png}
%	\includegraphics[height=0.17\textwidth]{figures/stir-water-2017.png}
%	\includegraphics[height=0.17\textwidth]{figures/spoon-peas-2016.png}
%    (C)~~~~~~~~~~~~~~~~~~~~(D)~~~~~~~~~~~~~~~~~(E)\\
%	\includegraphics[height=0.16\textwidth]{figures/screw-2016.png}
%    \includegraphics[height=0.16\textwidth]{figures/hammar-2017.png}
%	\includegraphics[height=0.15\textwidth]{figures/straw-2016.png}
%	\includegraphics[height=0.15\textwidth]{figures/towl-2017.png}
%	\includegraphics[height=0.16\textwidth]{figures/syringe-2016.png}
%	\includegraphics[height=0.16\textwidth]{figures/grasp_bnc.png}\\
% (F)~~~~~~~~~~~~~~~~~~~~~(G)~~~~~~~~~~~~~~~~~(H)~~~~~~~~(I)\\
%    \includegraphics[height=0.17\textwidth]{figures/grasp_sq_peg.png}
%	\includegraphics[height=0.17\textwidth]{figures/grasp_belt.png}
%	\includegraphics[height=0.17\textwidth]{figures/gasp_wire.png}
%    (J)~~~~~~~~~~~~~(K)~~~~~~~~~~~~~~~~~~~~~~~(L)~~~~~~~~~~~~\\
\caption{Tasks reflecting grasp challenges. Description of the tasks is provided in Table~\ref{tab:grasp-challenges}.}
	\label{fig:grasp-challenges}
\end{figure}

\subsubsection{Grasping with imperfect perception} 
Perception errors cause most failures in grasping. If the object's pose is estimated seriously wrong, the robot could knock over the object or push it away, end up with an unstable grasp, or completely miss the object. For end-to-end approaches, even if there is no separated pose estimation step, the noise in perception layers still dramatically affects the grasp results. Unfortunately, it is almost impossible to have ``perfect'' perception.  Noise can always find a way to creep in. So grasping approaches should be prepared to deal with perception uncertainties. In RGMC, most tasks require robots to handle a bad grasp caused by perception errors since many tasks have challenging objects.  

\subsubsection{Objects with challenging shapes and surfaces} 
Even with perfect vision, some objects can still be challenging to pick up or manipulate. It could be because there is a minimal good surface area for grasping. Curved and narrow surfaces are challenging to grasp. For example, the curved finger rings of a pair of scissors pose a significant challenge to a robotic hand (Fig. \ref{fig:grasp-challenges}a). Holding and manipulating scissors with them are both very difficult. 
Another challenging task is to pick up a saucer with a cup on it with one hand, since it is difficult to hold the edge of the saucer and balance the unknown torque generated by the cup's weight (Fig. \ref{fig:grasp-challenges}b).

\subsubsection{Grasping objects in clutter} 
Picking up an object in clutter is quite challenging. Its neighbor objects could block the robotic hand and pose planning difficulties, since usually a gripper would need to grasp on two sides of the object, and the hand approaches from another side.  One task that reflects this challenge is picking objects from a shopping basket and placing them in defined areas (Fig. \ref{fig:grasp-challenges}c). Grasping objects in clutter for manipulation is more challenging since the object would need to be held in a certain way to provide needed manipulability \cite{sun2016robotic} and interactive force \cite{lin2016task}. Therefore, the grasping points and orientations for manipulation are more limited than for picking. When objects are in clutter, a robotic hand may not be able to reach and grasp the object in the desired way. For example, in the stirring water task (Fig. \ref{fig:grasp-challenges}d), and pea-picking using a spoon task  (Fig. \ref{fig:grasp-challenges}e), the spoons are in a silverware organizer and they are difficult to pick up. 

\subsubsection{Regrasping} 
A robot may pick up an object in one grasp but may need to change to a different grasp for manipulation.  Then the robot would need to regrasp the object after it is picked up. As mentioned before, in both stirring and pea-picking tasks, the grasp of the spoon should be adjusted after it is picked up.  In the task of putting on or removing a bolt from a nut with a nut driver, since multiple turns are required, a robotic hand would need to regrasp the nut driver to overcome the rotation limit of the wrist (Fig. \ref{fig:grasp-challenges}f). Due to the number of fasteners, making regrasp operations efficient is challenging.

\begin{table}[]
    \centering
    \caption{Tasks reflecting grasp challenges}
    \label{tab:grasp-challenges}
    \begin{tabular}{p{0.13\textwidth}|p{0.3\textwidth}}
        \hline
        Research challenges & Tasks \\
        \hline\hline
        Objects with challenging shapes and surfaces
        & -Cut a piece of paper using a pair of scissors (Fig.~\ref{fig:grasp-challenges}a)\newline 
        -Transfer a cup on its saucer (Fig.~\ref{fig:grasp-challenges}b)\\
        \hline
        Grasp in clutter & -Pick and place objects from a shopping basket (Fig.~\ref{fig:grasp-challenges}c)\\
        & -Pick up silverware from a silverware organizer (Fig.~\ref{fig:perception-challenges}a)\\
     \hline
        Regrasp & -Assemble fasteners\\
        & -Use a spoon to stir water in a cup (Fig.~\ref{fig:grasp-challenges}d)\\
        & -Use a spoon to pick up peas (Fig.~\ref{fig:grasp-challenges}e)\\
        \hline
        Grasp for manipulation &  -Hammer a nail (Fig.~\ref{fig:grasp-challenges}g) \newline -Stir water with a spoon (Fig.~\ref{fig:grasp-challenges}d)\\
        & -Assemble/disassemble electrical connectors (Fig.~\ref{fig:grasp-challenges}i)\\  
        & -Insert pegs (Fig.~\ref{fig:grasp-challenges}j)\\
        & -Insert/screw fasteners (Fig.~\ref{fig:grasp-challenges}f)\\
        & -Insert/mesh gears
        (Fig.~\ref{fig:manipulation-challenges}b)\\
        & -Insert/rout belts/wires (Fig.~\ref{fig:grasp-challenges}k,l)\\
        
        \hline
        Grasp for in-hand manipulation & -Fully extend and fully press syringe (Fig.~\ref{fig:grasp-challenges}h) \newline 
        -Grasp and use scissors to cut a piece of paper to half along a line (Fig.~\ref{fig:grasp-challenges}a)\\        
        & -Pick up ice cubes using tongs (Fig.~\ref{fig:perception-challenges}d)\\  
        \hline
        \end{tabular}
\end{table}

\subsubsection{Grasping for manipulation} 
For some manipulations, an object should be grasped in a certain way so that the manipulation can be performed efficiently and the object has less chance of being dropped during the operation \cite{lin2015task, lin2015grasp}. We have designed nail hammering (Fig.~\ref{fig:grasp-challenges}g) and water stirring tasks to gauge if the team considers the manipulation requirements  in grasp planning. Performing force-based manipulation for insertions requires adequate grasping force to avoid movement of the object within grasp and ultimate insertion failure (Fig.~\ref{fig:grasp-challenges}j). In addition, grasping non-rigid objects to effectively control their shape, as in threading belts on pulleys (Fig. \ref{fig:grasp-challenges}k) and  wire routing (Fig. \ref{fig:grasp-challenges}l), is a very challenging problem.  

\subsubsection{Grasping for in-hand manipulation} 
In-hand manipulation is still a very challenging research area. We have designed several tasks that would require some in-hand manipulation after grasping. These include extending and pressing a syringe (Fig. \ref{fig:grasp-challenges}h), cutting paper with a pair of scissors (Fig. \ref{fig:grasp-challenges}a), and using tongs to pick up ice cubes (Fig. \ref{fig:perception-challenges}d), but they have not been satisfactorily solved so far.

Table \ref{tab:grasp-challenges} summarizes the tested grasp challenges and the tasks reflecting them in the RGMCs. 

\begin{figure}
	\centering
	\includegraphics[width=\columnwidth]{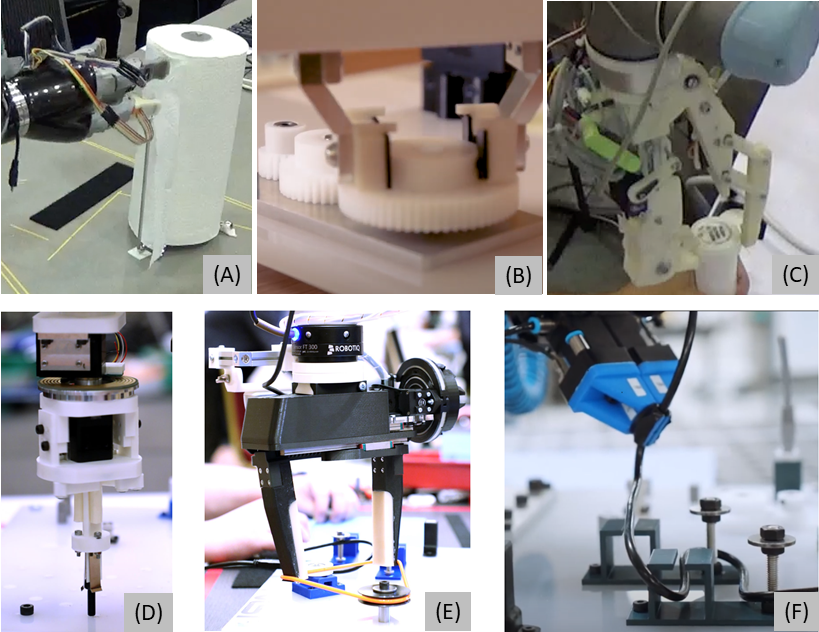}
	%\includegraphics[height=0.18\textwidth]{figures/towl-2017.png}
	%\includegraphics[height=0.18\textwidth]{figures/pour-2017.png}
	%\includegraphics[height=0.18\textwidth]{figures/lock-cap-2017.png}
	%    (A)~~~~~~~~~~~~~~~~~~~~(B)~~~~~~~~~~~~~~~~~~~(C)\\
	%\includegraphics[height=0.205\textwidth]{figures/manip_screw.png}
	%\includegraphics[height=0.205\textwidth]{figures/manip_belt.png}
	%\includegraphics[height=0.205\textwidth]{figures/manip_wire.png}
	%    (D)~~~~~~~~~~~~~~~~~~~~(E)~~~~~~~~~~~~~~~~~~~(F)\\    
\caption{Tasks reflecting manipulation challenges. Description of the tasks is provided in Table~\ref{tab:manipulation-challenges}.}
	\label{fig:manipulation-challenges}
\end{figure}

\subsection{Research Challenges in Manipulation}

Manipulation motions can usually be generated with motion planning and motion control. In general, the challenges in manipulation are caused by imperfect perception and the lack of capability to predict a motion's outcome. Both can be extremely challenging.  In RGMCs, since the objects are provided to the teams beforehand (at least for several hours), they can model the objects and their behaviors to a certain extent, which dramatically reduces the challenges.  However, several tasks remain challenging because either the tolerance is tight or the behaviors are difficult to model. For instance, peg-in-hole tasks and tightening or loosening bolts (Fig.~\ref{fig:manipulation-challenges}d) require purposeful  motions.
Tearing up a paper towel (Fig.~\ref{fig:manipulation-challenges}a) and pouring water (Fig. \ref{fig:manipulation-challenges}b) require the robot to predict the kitchen roll motion and water flowing speed in response to a motion. Opening a bottle with a locking safety cap (Fig. \ref{fig:manipulation-challenges}c)  is a task that requests the robot to predict a pressing outcome. Since it is repeatable, it is easier to model than water and cloth. Assembling flexible belts and routing wires, while difficult, becomes more achievable with practiced routines and the use of CAD data. Table \ref{tab:manipulation-challenges} summarizes the tasks reflecting those challenges in the RGMCs. 

\begin{table}
    \centering
        \caption{Tasks reflecting manipulation challenges}
    \label{tab:manipulation-challenges}
    \begin{tabular}{p{0.13\textwidth}|p{0.3\textwidth}}
     \hline
     Research challenges & Tasks \\
      \hline  \hline
      Manipulation with imperfect perception & -Solve peg in hole with a wood board \newline -Assemble/disassemble electrical connectors \\ 
        & -Insert pegs (Fig.~\ref{fig:grasp-challenges}j)\\
        & -Insert/screw fasteners (Fig.~\ref{fig:manipulation-challenges}d)\\
        & -Insert/mesh gears
        (Fig.~\ref{fig:manipulation-challenges}b)\\
      \hline
     Manipulation that needs accurate prediction   & -Open a water bottle cap and pour water into a to-go cup \newline 
      -Pour water from a pitcher into a cup
      (Fig.~\ref{fig:perception-challenges}e)\newline
     -Tear a paper towel (Fig.~\ref{fig:manipulation-challenges}a)\\
     & -Open a bottle with a locking safety cap (Fig.~\ref{fig:manipulation-challenges}c)\\
     & -Threading flexible belts (Fig.~\ref{fig:manipulation-challenges}e)\\
     & -Routing wires
     (Fig.~\ref{fig:manipulation-challenges}f)\\
          \hline
        \end{tabular}
\end{table}

\begin{comment}

\subsection{Limitations of mechanical designs}

\subsubsection{Size and shape}
Deal with objects that are designed to be used by human hands.  
- handles
- pitcher handles
- cup handles
- saucer
size:
- compartment of a tray

\subsubsection{Dexterity}
The objects are designed to rely on human hand dexterity. 
- syringe
- scissors

\subsubsection{Softness}
surfaces are hard since human hands are soft. Soft contact provide large contact, large friction, and torque. 
It is difficult to be precise and soft. 

\subsubsection{Strength}
Enough strength and dexterity. A hand can hold hammer a nail, can also use scissors and syringe. 
- A arm can hold a pitcher with a liter of water. 
- hammer
- saucer
- locking safety cap
- finger strength of using scissor, syringe, and tongs. 
- twist open cap
A thin and soft object is also difficult to grasp and hold on. 
Tear away one piece of paper towel 

\subsubsection{Sensing}
Touch a holding a piece of paper towel, and 

\end{comment}

%%%%%%%%%%%%%%%%%%%%%%%%%%%%%%%%%%%%%%%%%%%%%%%%%%%%%%%%%%%%%%%%%
\section{Results and Progress}
Each year, seven to ten teams passed the initial screening and participated in the competition.  Teams are from world-renowned research groups in universities, prominent startups, and large corporations.  We can see that their improvement in performances aligns with the progress made by the robotics research community over the same period of time. Though their performance might not represent the absolute best solutions available at the time, their successes and struggles in the RGMCs reflect research progress and remaining challenges. 

\subsection{Overview}
In the four RGMCs, twenty-three tasks have been used to measure the contestant's performance in the service track, some used in multiple years with stricter requirements toward achieving real-world application.  In the beginning, in 2016, the teams were allowed to predefine the locations of the objects with little to no randomness. Then in 2017 and afterward, the teams were only allowed to define object regions based on the robot's workspace and the organizers randomly placed the objects in those regions.  In 2016 and 2017, only the best possible result for every task was counted, while in 2019 and 2020, the total score of five trials was counted. In 2019, the total allowed time was 90 minutes, while in 2020, the total allowed time was 60 minutes. The manufacturing track was first introduced in 2017, and teams were allowed to set the position of the task board and kit layout, and optional random placement provided bonus points.  The dominant solution was the use of lead-through programming.  To encourage the use of perception, in 2019 and 2020 randomized placement was required and CAD data was supplied as an option for extracting locations of assembly components on both the task board and kit layout relative to a coordinate frame. While the 2017 task board contained only insertion and fastening operations, the 2019 and 2020 task boards introduced wire routing and belt on pulley operations.
%The supplemental document\footnote{\url{https://rpal.cse.usf.edu/competition_iros2021/Supplemental-RA-L-RGMC.pdf}} provides details of the tasks, the best scores, unsolved tasks, and performances of the RHGM teams based on the data and organizers' observation. 
The Appendix provides details of the tasks, the best scores, unsolved tasks, and performances of the RHGM teams based on the data and organizers' observation. 

\vspace{-2mm}
\subsection{Research Progress and Challenges in Perception}
\subsubsection{Objects with shiny surfaces} 
In 2016, teams were struggling with silverware due to their shiny surfaces. In 2020, several teams could reliably segment a shiny spoon from other spoons and estimate its pose for grasping. They managed the reflection by controlling the lighting using multi-flash, similar to the technique in \cite{agrawal2010vision}. Objects with shiny surfaces may not be a significant challenge if the object model is known and its location variation is limited. 

\subsubsection{Translucent or transparent objects}
Translucent objects still pose a significant challenge to perception, especially for randomly stacked objects such as ice cubes in a bucket. No team was able to finish the picking ice cube task. For the pouring task, all teams completed it using hard-coded pouring motions. We do not think any team could estimate the water level in the receiving cup when the water is transparent. Even though several promising pouring approaches are available, none of the teams used them. So, when an object is truly transparent, alternative perception other than vision could be used.  On the other hand, randomly stacked translucent objects are still very challenging. 

\subsubsection{Tasks requiring high precision or accuracy}
In plugging a USB light into a socket task, several teams were able to finish the task. Some teams are slower than others.  But in general, the peg-in-hole problem is not a significant challenge if both models are known.  

%\paragraph{Other challenges.} If the objects are unknown (no learning data or models), the difficulty level  Segmentation, 

%%%%%%%%%%%%%%%%%%%%%%%%%%%%%%%%%%%%%%%%%%%%%%%%%
\vspace{-2mm}
\subsection{Research Progress and Challenges in Grasping}
\subsubsection{Grasp with imperfect perception} 
Teams in RGMCs have gradually incorporated approaches to handle perception uncertainties, since they found that it was one of several major reasons that make their solutions slow and unreliable. It is a significant challenge that requires both research and engineering work. 

\subsubsection{Objects with challenging shapes and surfaces} 
Teams have explored several options and made good progress. Several teams tried an automatic tool-change approach that allows them to swap grippers. Several grippers were designed and made to deal with challenging shapes and surfaces. Some others  developed a gripper that incorporates multiple fixtures. They can flip or rotate so that the right fixture is in contact with the object. Many smart and inspiring designs have come out, but they are tailored and calibrated on known objects. It is a significant challenge for cases with unknown objects or under unseen situations. 

\subsubsection{Grasp objects in clutter} 
RGMCs only have several tasks with objects in clutter, and the level of cluttering is moderate. On the one hand, we observed the teams made progress over the years in dealing with closely lying objects. But on the other hand, we also observed that cluttered environments still slowed down their performance dramatically. So it is still a challenge, and we would expect it is more difficult if the objects are unknown and the situations are new. 

\subsubsection{Regrasp} 
Many tasks require the robot to adjust the grasp after picking up an object. We have observed many teams put down the object and used the setup to position or orient the object in a certain way and then regrasped it. Many of the pick-place-regrasp routines are well crafted and impressive.  However, teams have been avoiding in-air regrasp even it would be more efficient. Adjusting a grasp or regrasping in the air is still very challenging.  

\subsubsection{Grasp for manipulation} 
Several tasks require grasp planning while considering manipulation requirements.  We have mostly seen that teams predefine several grasp points based on their experiences and simulations.  As far as we know, teams have avoided computing and searching for proper grasp points on the fly. The approach would fall apart if the objects are unseen or different from the provided models.  For known objects, the predefined grasp point approach seems sufficient. 

\subsubsection{Grasp for in-hand manipulation}
In-hand manipulation remains the biggest challenge, where teams have been unable to complete tasks that utilize scissors, syringes, and tongs. These tasks, in most cases have been removed from RGMC due to their high difficulty.

\vspace{-2mm}
\subsection{Research Progress and Challenges in Manipulation}
\subsubsection{Manipulation with imperfect perception}
Many teams have no problem dealing with tasks testing manipulation with imperfect perception. In general, the available object models and setups and tactile/force sensors allow teams to complete those tasks successfully.  
\subsubsection{Manipulation that needs accurate prediction} 
Teams did little in predicting the outcomes of a manipulation action. Most manipulation motions were generated based on obstacle avoidance. As far as we know, they have avoided modeling dynamics. They usually created a list of possible outcomes and matched a set of predefined motion strategies to those outcomes. This approach is not sufficient when dealing with flexible objects and fluids, since a finite enumeration of the states of those objects is usually not feasible. Even though the teams have not provided satisfactory solutions, we believe that recent research progress could be used to solve some of the challenges \cite{chen2019accurate, huang2021robot}.  Table \ref{tab:perception-progress} summarizes the tested challenges and the progress in the RGMCs. 

\begin{table}[h!]
    \centering
    \caption{Perception challenges and progress}
    \label{tab:perception-progress}
    \begin{tabular}{p{0.13\textwidth}|p{0.3\textwidth}}
        \hline
        Research challenges &Progress \\
        \hline \hline
        Shiny objects & Mostly solved for standalone objects if their models are known. \\
            \hline
        Translucent objects & Still very challenging for randomly stacked translucent objects. \\
            \hline
        Requiring high precision or accuracy & Largely manageable if the objects are known. Implementations are incorporating force control and CAD data.\\  
            \hline
        Grasp with imperfect perception & It is the major reason that solutions are slow and unreliable. It is a significant challenge that  requires research and a lot of engineering work. \\
        \hline
        Objects with challenging shapes and surfaces & It remains a significant challenge even though special-purpose grippers and fingers are designed. \\
        \hline
        Grasp objects in clutter & It is still a challenge, and we would expect it is more difficult if the objects are unknown and the situations are new. \\
        \hline
        Regrasp & Many teams developed impressive pick-place-regrasp routines, but adjusting a grasp or regrasping in the air is still very challenging.\\
        \hline
        Grasp for manipulation & For manipulating known objects in a known condition, the predefined grasp point approach seems to work well. However, this approach would fall apart if the objects are unseen or the conditions are new. \\
        \hline
        Grasp for in-hand manipulation & In-hand manipulation remains the most difficult challenge, as it requires accurate prediction of the motion effects and suitable control of the end effector.\\
        \hline
        Manipulation with imperfect perception & The available object models, design data, and tactile/force sensors allow teams to complete those tasks successfully. \\
        \hline
        Manipulation that needs accurate prediction & Teams have avoided modeling dynamics for prediction, even though recent research progress could be used to solve some of the challenges.\\
        \hline
        \end{tabular}

\end{table}

%\subsubsection{Progress by year}
%In the 2016 RGMC, we have a special track called hand-in-hand track to evaluate the capabilities of robotic hands.  Each team in the track provided a robotic hand without a robotic arm to a volunteer.  Then the volunteer holds the robotic hand to perform the same ten tasks in the fully-autonomous track. The team controlled the finger/gripper of the hand through a computer.  The volunteer and the teams had one day to practice before the competition.  Two teams were able to finish all 10 tasks in about 20 to 30 minutes.  One gripper was a two gripper hand \cite{} and the other was a five finger hand with one-DOF control \cite{}. The results show the capability of a combined hand-robot system. We have observed that volunteers have learned to use the robotic hands as tools and use their own perception and intelligence to ???. In the autonomous track, teams were allowed to try the tasks multiple times within 120 minutes. 

%%%%%%%%%%%%%%%%%%%%%%%%%%%%%%%%%%%%%%%%%%%%%%%%%
\vspace{-2mm}
\section{Discussion and Future Directions}
%how can we summarize the results of previous years? 
%How can we analyze the evolution of the outcome of the competition? 
%Have teams evolved over time, are they closer to solve in a satisfactory manner the tasks we proposed?

This paper provided an overview of the tasks and challenges proposed in the RGMCs. 
Different from many evaluation setups in the literature, competitions such as RGMC usually require participants to run demos at a certain time and to complete tasks in a limited amount of time. In addition, participants usually do not have total control of the setup, making the competition setup realistic to real-world applications. 

Over the years, we have seen that tasks that were initially deemed as hard, have been solved using different ingenious approaches. 
Especially in the first editions, teams tried to solve the most difficult challenges with highly-engineered solutions. But with progress made in research, we have seen less and less hard-coded routines, less use of predefined grasp points, which have led to solutions becoming more effective and reliable. 
For software, the most significant challenges identified by the teams are related to perception. The arrival of learning-based approaches has facilitated solving tasks that use known objects, as they can be readily available for system training. The generality of such approaches to be applicable to familiar and fully unknown objects is still a large challenge.
Learning-based approaches have been also lately used for defining grasping configurations, thus avoiding the reliance on predefined grasp poses.
For motion planning, most teams relied on open libraries, such as the Open Motion Planning Library (OMPL) \cite{ompl} or MoveIt \cite{moveit}, and using Robot Operating System (ROS) \cite{ros} in most cases as middleware. These openly-available resources enable teams to more efficiently create solutions.

We saw that some of the RGMC tasks still represent a significant challenge. The most clear example are tasks requiring some degree of in-hand manipulation, which were effectively removed from the last RGMC editions. This reflects the state of industry, where most end-effectors are two-finger grippers that are used for pick-and-place tasks. Industrial applications usually prefer to design special fingertips for grasping different types of objects, or using a robot accompanied by a  tool changer that endows the robot with the ability to switch end-effectors for performing multiple tasks. The promised generic dexterity of multi-finger end-effectors still seems more a research topic rather than a viable commercial solution, mainly due to the higher mechatronic and control complexity and higher cost associated with multi-finger grippers. 

There has been a significant increase in manufacturing track solutions that utilize CAD data. Typically, perception or force solutions are used to localize the task board with subsequent localization of part assembly points on the task board using CAD data.  Improvements to the CAD pipeline could result in improved data formats to transmit dimensions, geometric features, relative part positions, mating descriptions, and tolerances from CAD systems. In addition, the inclusion of force-based assembly parameters within a CAD system, which could be best specified by a designer who is most familiar with the mechanical properties of the parts to be assembled, should be considered. Methods for automatic robot program generation using this data should also be considered.

In terms of robotic devices, participants have used a full range, from experimental robotic hands and arms to off-the-shelf robot arms and simple parallel grippers. Hardware for perception is nowadays relatively standard, with depth cameras dominating the landscape, although different teams have been using other types of sensors (stereo cameras, laser-based perception, in-hand cameras). Integration of different hardware and software components remains, however, a large challenge, requiring hours of dedication to reach a stable and robust execution during the demonstrations. In most RGMC editions, teams had to transport their own equipment to the competition locations. The pandemic required the 2020 edition to be a fully online competition, which greatly alleviated the logistics for participation of teams around the world. Moving toward a standard remote lab for testing different approaches for solving a set of tasks seems like a reasonable approach to enhance comparability and reproducibility of results in robotics in the near future.

New robot designs provide improved methods for fast lead-through programming using direct interactions between the operator and the robot. Teams leveraging these solutions typically score well by programming the most difficult high scoring tasks first, to accumulate as many points as possible within a given time frame.  Competition format changes to discourage solutions that only use lead-through programming could include: a required part assembly order, less tasks per board, and unknown task board offset and part variations.  

Several successes have ensued RGMC teams. Dorabot \cite{dorabot}, for instance, participated in the 2016 and 2017 RGMC competitions with prototypes of their dexterous modular robotic three-fingered hand, which is now a product.  Focused on robotic logistic solutions, the company now has products including a five-fingered hand, and mobile manipulators.  Robotic Materials \cite{rm} emerged as a startup out of the University of Boulder following the RGMC 2016/2017 competitions, and participated in RGMC 2019/2020 competition using their prototype smart gripper, which integrated position, force, and depth camera sensing modalities.  Their work on this gripping system continues with the development of easy-to-use programming interfaces aimed at small- and medium-sized manufacturers. Southern Denmark University \cite{usd} research focuses on the development of a flexible workstation for automated assembly tasks that can be used commercially.  Their generalized solution is proving to be robust and streamlined, achieving the first ever perfect score in the 2020 RGMC Manufacturing Track.

\section*{Acknowledgment}

\vspace{-1mm}
We would like to acknowledge all RGMC co-organizers.  In addition to the authors, they are Nadia Cheng (2016), Hyouk Ryeol Choi (2016), Zoe Doulgeri (2016, 2017), Erik D. Engeberg (2016, 2017), Kris Hauser (2016), 
Nancy Pollard (2016), Zeyang Xia (2016), Yasuyoshi Yokokohji (2017, 2019), Zoe Doulgeri (2017), Yunjiang Lou (2017), Hyungpil Moon (2017), Juxi Leitner (2019), Rong Xiong (2019), and Adam Norton (2020).
We also acknowledge Kenneth Kimble of NIST for numerous detailed competition task board designs.

\begin{comment}
We want to thank the IROS organizers and especially the competition co-chair for provide space, infrastructure and logistics support. They are ???

We also want to thank all sponsors who have helped in paying travel cost of teams, setup cost, and awards. They are
\begin{itemize}
    \item IROS, 2016 and 2017, setup cost and travel support
    \item ROBOTOUS, 2016, one Five 6-axis F/T sensors (value of \$1,000) and \$1,000 cash;
    \item OptoForce, 2016 and 2017, two 6-axis F/T sensors (value of 2,500 EUR,
    \item RightHand Robotics, One TakkTile Starter Kit (value of \$300), and 6 months loan of the ReFlex TakkTile robotic gripper (value of \$10,000), 
    \item World Robot Summit, 2017, \$30,000?
    \item Cainiao Smart Logistics Network, 2019, \$??
\end{itemize}

\end{comment}

\bibliographystyle{IEEEtran}
\vspace{-2mm}
\bibliography{ref}
\clearpage 
\newpage
\newpage
\newpage


\section*{Supplementary material}

# Appendix

## 1. Tasks Description

A. *Service Track*
Twenty-three tasks have been used in the competitions.

| Task | Years | Difficulty at the time | Common tricks | Best Performance (0-1) |
|---|---|---|---|---|
| Use a spoon to pick up peas | 2016 | | Spoon was placed in a cup for pickup | 1 |
| Pick up a fork, a knife, and a spoon from a silverware tray | 2017 | Try slots are smart and the silverwares are shiny | | 0.66 |
| Pour water from a pitcher | 2017 | Pitcher with water is too heavy | | 0 |
| Use a spoon to stir water in a cup | 2016, 2017 | | Rotate spoon back and forth | 1, 1 |
| Use scissors to cut a piece of paper to half along a line | 2016 | Difficult to grasp scissors and manipulate it | | 0 |
| Fully extend syringe and then fully press syringe | 2016 | Cannot extend | Press against the table surface | 0.5 |
| Putting on or removing bolts from nuts with a nut driver | 2016 | | Nut driver stays with the bolt and won't fall. It allows the hand to fully release and regrasp. | 1 |
| Shake out salt from a salt shaker to a defined location | 2016 | | Preprogram a routine | 1 |
| Grasp a towel on the table and hang it onto a support | 2016 | Put it on the rack is challenging | The rack has spikes. Dropping the towel at a certain point allows the towel | 1 |

| | | | to fall on the rack. | |
|---|---|---|---|---|
| Open a bottle with a locking safety cap | 2017 | Pressing and then turning. | Preprogram a press-and-turn is good enough. | 1 |
| Hammer nails | 2016, 2017 | Holding on to the hammer and swing is difficult. Precision is also challenging. | Holding the hammerhead and pressing on the nails. Since it is a foam board, that pressure is sufficient for the nails to go into the foam board. | 0, 1 |
| Peg in hole insertion tasks with a shape board | 2017 | Alignment challenge | Teams can try again and again. | 0.83 |
| Grasp and cleanly tear away one piece of paper towel | 2017 | Holding on the paper is difficult while tearing | | 0.5 |
| Grasp a plug and insert it into a socket | 2016, 2017 | | Slow, but works. | 1 |
| Transfer a cup on its saucer | 2017 | Saucer has little area to hold on to. | Special gripper works fine. | 1 |
| Take the top to-go cup from the cup pile and place it on the table | 2019, 2020 | Plastic cups are shiny and translucent. | Some teams prepared an upside-down setup. It is easier. | 1, 1 |
| Put one teaspoon of matcha green tea powder into the to-go cup | 2019, 2020 | Scoop the right among and not drop any in transporting. | Pick up a spoon, place it against the tray so the hand can regrasp it, and hold it in a preferred orientation. | 0.63, 1 |
| Open the water bottle cap and pour water into the to-go cup | 2019, 2020 | Pouring the desired amount from a bottle with a random water level. | Caps were pre-opened (allowed). Pouring was prescribed based on time. Pouring | 0.64, 0.8 |

| | | | three cups from three full bottles removed randomness. | |
|---|---|---|---|---|
| Use the teaspoon to stir water | 2019, 2020 | Stop at the desired state. | The liquid becomes blended after a certain number of stirs. | 1,1 |
| Pick up two ice cubes and drop them into the to-go cup | 2019, 2020 | Pick up tongs are shiny, ice Cubes are translucent. | Can pick up tongs. Put the tongs randomly into the bucket may result in some success in random. | 0.2, 0.33 |
| Put a lid on the to-go cup | 2019, 2020 | Pressing the lid till it locks. | Special gripper for pressing | 0.33, 0.89 |
| Transfer straw into a to-go cup with lid | 2016, 2017, 2019, 2020 | Pick up one straw from many straws. Insert into the cross at the center of the lid. | The lid will deform a little bit to tolerate the inserting error. | 0.5, 1, 1, 0.5 |
| Place the to-go cup to a defined location | 2019, 2020 | | | 1, 1 |

\* Best Performance column lists the best score percentages for the corresponding year (the higher the better). For example, on the "Use a spoon to stir water in a cup" row, "1, 1" indicates that the best scores of the task in both 2016 and 2017 were 100%.


***Challenging or Unsolved Tasks***

The following tasks remain very challenging and robots still have trouble performing them perfectly and consistently:
- Pick up two ice cubes and drop them into the to-go cup
- Open the water bottle cap and pour water into the to-go cup
- Use scissors to cut a piece of paper to half along a line
- Hammer nails
- Pick up a fork, a knife, and a spoon from a silverware tray
- Fully extend syringe and the full press syringe
- Peg in hole insertion tasks
- Grasp and cleanly tear away one piece of paper towel
- Put a lid on the to-go cup

- Transfer straw into a to-go cup with lid

B. *Manufacturing Track*
Twenty-five tasks have been used in the competitions, both for assembly and disassembly operations.

| Task Class | Years | Description | Best Performance* (0-1) |
|---|---|---|---|
| Threading Fasteners | 2017, 2019, 2020 | Thread screw/nut and drive until seated. | A: 0.5, 0.96 ,1<br>D: 0.25, 0.33 ,1 |
| Insertions | 2017, 2019, 2020 | Insert peg or connector until fully seated | A: 0.97,0.76 ,1<br>D: 1,0.89 ,1 |
| Wire Routing (deformable object) | 2019, 2020 | Route wire through guides | A: 0.5, 1<br>D:1, 1 |
| Belt on Pulley (deformable object) | 2019, 2020 | Insert belt onto two pulleys and tighten | A: 0.67, 1<br>D: 1, 1 |

*A - Assembly   D - Disassembly*

**Challenges**
Threading Fasteners - many teams use valuable time in trying to utilize a robot's flange joint to drive fasteners.  This is more evident as the number of fastener operations increased from 4 (2017) to 18 (2019 & 2020). Teams performing best in this category utilize an external rotary tool to accomplish driving fasteners.

Insertions are a relatively successful task among teams, with the major problems being the BNC connector which requires a three-step procedure of insert - twist - lock.  Organizers expect to introduce tighter tolerances and more snap-to-lock operations in future competitions.  Current tolerances on pegs are quite loose.  Most teams have incorporated force sensing into their solutions to support insertions.

Wire routing is still a difficult task for most teams.  Teams succeeding in this operation utilize CAD data to locate routing positions and incorporate back and forth (wiggle) motion routines to fit the wire into the groove.  The fastest solution to date incorporated two robots where one supported the cable length while the other worked the segment being routed with a wiggle motion.

Belt on Pulley task has been solved by many teams using both one and two manipulator systems. It is observed that in a one manipulator system, the solution requires more sophisticated path planning and gripper finger design plays a larger role.

It is difficult to measure progress as the task boards grow in complexity from competition to competition. This is primarily due to the continued development of primary task boards (see: https://www.nist.gov/el/intelligent-systems-division-73500/robotic-grasping-and-manipulation-assembly/assembly) where competition task boards contain a subset of operations from each of the primaries. The intent is to use competitions to raise awareness to the range of assembly task boards in hopes that they will be used in more detailed experiments to benchmark progressing research work. Teams also try new solutions in subsequent competitions and often perform more poorly due to technical difficulties during these one-shot competitions, indicating that reliability is an important measure and multiple runs should be used when using these benchmarks to measure research results.

## 2. General Differences Among the Years

*Year 2016 (Daejeon, Korea)*
- *Th*e competition this year had three tracks.
  - Track 1: Hand in hand track was designed to explore the capability of the robotic hands.
  - Track 2: Autonomous track, where the locations of the objects are predefined with little to no randomness.
  - Track 3: Simulation
- The teams were permitted to accomplish multiple runs in a two-hour time limit.
- Objects were provided to the teams a day before the competition.

The detailed description of the rules in 2016 is at
http://www.rhgm.org/activities/competition_iros2016/

*Year 2017 (Vancouver, Canada) -first manufacturing track*

*Ser*vice track:
- More randomness was introduced.
- The teams can define the regions based on their robot's workspace.
- The organizers randomly place the objects in those regions.
- The teams can try multiple times within two hours.
- Objects were provided to the teams a day before the competition.

Manufacturing track: (includes World Robot Summit pretest gear assembly)
- Assembly/Disassembly of task board containing peg insertions, gear meshing, electrical connector insertions, and nut threading
- WRS - assembly of two spur gears, shafts, spacers, and nuts to a baseplate

- Teams defined the location of the base part relative to the robot workspace
- Objects were made available to the teams many weeks prior to the competition.
- Tier 1 & tier 2 bonus points for use of marker-based or no marker-based perception (no lead through programming)
- Taskboard disassemble/assemble (20 min./ 40 min.) - gear assembly (60 min.)

The detailed description of the rules in 2017 is at
http://www.rhgm.org/activities/competition_iros2017/

*Year 2019 (Macau, China)*

Service track:
- The teams defined the regions based on their robot's workspace.
- The organizers randomly place the objects in those regions.
- The teams were required to complete the cycle of tasks (each cycle contains eight tasks) five times within 90 minutes.
- Objects were provided to the teams a day before the competition.

Manufacturing track:
- Computer Aided Design - CAD data introduced
- Assembly/Disassembly of task board containing peg insertions, gear meshing, electrical connector insertions, bolt threading, wire routing, belt on pulleys
- The organizers randomly place the task board within the team workspace
- Teams were provided with a practice task board, kit layout, parts for assembly/disassembly, and CAD data.
- New task board, kit layout, and components with different task board and kit layout component locations for actual competition. Another set of the same parts provided.
- Subtask 1 - disassembly (40 minutes / 92 points) and subtask 2 - assembly (80 minutes 181 points)

The detailed description of the rules in 2019 is at
https://rpal.cse.usf.edu/competition_iros2019/

*Year 2020 (online competition)*

Service track:
- The teams purchased the items in the competition and knew all the objects beforehand.
- Through video conferencing, the organizers asked the teams to manually move the objects around a little bit to introduce randomness.
- The teams have little time for error recovery since they were asked to do the whole thing (containing eight tasks) five times in one hour.

Manufacturing track:
- Same task board design as 2019; however, fastener operation in aluminum instead of plastic for increased thread durability
- Teams were provided with a practice task board, kit layout, components, and CAD data.

- New task board, kit layout and components with different task board and kit layout component locations for actual competition. Separate package opened by remote teams at the start of the trial.
- Teams were required to randomly place task board and kit layout within the robot system workspace prior to the start of subtask 1.
- Subtask 1 - disassembly (40 minutes / 92 points) and subtask 2 - assembly (80 minutes 181 points).

The detailed description of the rules in 2020 is at
https://rpal.cse.usf.edu/competition_iros2020/


*Best scores across the years*

| Best score per track | 2016 | 2017 | 2019 | 2020 |
| --- | --- | --- | --- | --- |
| Best Score in Service Track | 165/200 | 148/235 | 515/1000 | 680/1000 |
| Best Score in Manufacturing Track | - | 169/180 | 113/273 | 317/273* |

*Team SDU achieved a perfect score and was awarded a time bonus for finishing before the allotted competition time period. While team SDU only competed in RGMC 2020, they competed and were a top-scoring team in the 2018 World Robot Summit: Assembly Challenge.


### 3. Participating teams

Year 2016. Total of ten teams participated. Seven teams participated in Hand-in-hand track, Five participated in the Autonomous track, Four teams participated in the Simulation track. Some teams participated in multiple tracks.
Year 2017. Total of seven teams participated. Seven teams participated in the Service track, Six participated in the Manufacturing track. Some teams participated in multiple tracks.
Year 2019. Total of ten teams participated. Seven teams participated in the Service track; five teams in the manufacturing track, and five teams in the logistics track. Some teams participated in multiple tracks.
Year 2020. Total of eight teams participated. Four teams participated in the Service track, four teams participated in the manufacturing track.

Eight teams participated for more than one year. They are:
- University of Colorado at Boulder 2016, 2017, 2019, 2020
- Tsinghua University, 2016, 2017, 2019
- Cothink Robotics, 2017, 2019, 2020

- CambridgeARM, UK, 2017, 2019
- FEIFAN AI, 2017, 2019
- SIAT, 2019, 2020
- New Dexterity, 2019, 2020
- JAKS, 2019, 2020

Team scores in the Service Track

| Teams | 2016 | 2017 | 2019 | 2020 |
|---|---|---|---|---|
| Univ. Colorado at Boulder | 100 | 90 | | |
| Tsinghua University | 165 | 113 | 435 | |
| Cothink Robotics/ | | 135 | 515 | 680 |
| CambridgeARM | | 148 | 509 | |
| FEIFAN AISASIA | | | 430 | 675 |
| SIAT | | | 331 | 622 |
| JAKS | | | 183 | 45 |

Team scores in the Manufacturing Track

| Teams | 2017 | 2019 | 2020 |
|---|---|---|---|
| University of Colorado at Boulder / Robotic Materials | 134 | 104 | 27 |
| FEIFAN AI | 169 | 33 | |
| JAKS | | 42 | 18 |
| New Dexterity | | 113 | 44 |

**Technology of Winning Teams**

Winning teams have demonstrated their solutions in the Youtube Channel of Robotic Grasping and Manipulation Competition

https://www.youtube.com/channel/UC8F4S4EjTAo2JbwdzbyKnCA


\end{document}